\documentclass{article}

\PassOptionsToPackage{round}{natbib}
\usepackage[final,main]{neurips_2025}

\usepackage[utf8]{inputenc}
\usepackage[T1]{fontenc}
\usepackage{hyperref}
\usepackage{url}
\usepackage{booktabs}
\usepackage{amsfonts}
\usepackage{amsmath}
\usepackage{amssymb}
\usepackage{nicefrac}
\usepackage{microtype}
\usepackage{xcolor}
\usepackage{graphicx}
\usepackage{subcaption}
\usepackage{multirow}
\usepackage{float}
\usepackage{placeins}

\usepackage{etoolbox}

\graphicspath{{figures/}}

\makeatletter
\renewcommand{\@trackname}{}
\makeatother

\title{Dynamic Spatio-Temporal Graph Neural Network for Early Detection of Pornography Addiction in Adolescents Based on Electroencephalogram Signals}

\author{%
  \textbf{Achmad Ardani Prasha}$^{1}$\thanks{Corresponding author: 41523010005@student.mercubuana.ac.id} \quad
  \textbf{Clavino Ourizqi Rachmadi}$^{1}$ \quad
  \textbf{Sabrina Laila Mutiara}$^{1}$ \\
  \textbf{Hilman Syachr Ramadhan}$^{1}$ \quad
  \textbf{Chareyl Reinalyta Borneo}$^{2}$ \quad
  \textbf{Saruni Dwiasnati}$^{1}$ \\[0.5em]
  $^{1}$Faculty of Computer Science, Universitas Mercu Buana, West Jakarta, Indonesia \\
  $^{2}$Faculty of Psychology, Universitas Mercu Buana, West Jakarta, Indonesia
}

\begin{document}

\maketitle

\begin{abstract}
Adolescent pornography addiction requires early detection based on objective neurobiological biomarkers because self-report is prone to subjective bias due to social stigma. Conventional machine learning has not been able to model dynamic functional connectivity of the brain that fluctuates temporally during addictive stimulus exposure. This study proposes a state-of-the-art Dynamic Spatio-Temporal Graph Neural Network (DST-GNN) that integrates Phase Lag Index (PLI)-based Graph Attention Network (GAT) for spatial modeling and Bidirectional Gated Recurrent Unit (BiGRU) for temporal dynamics. The dataset consists of 14 adolescents (7 addicted, 7 healthy) with 19-channel EEG across 9 experimental conditions. Leave-One-Subject-Out Cross Validation (LOSO-CV) evaluation shows F1-Score of 71.00\%$\pm$12.10\% and recall of 85.71\%, a 104\% improvement compared to baseline. Ablation study confirms temporal contribution of 21\% and PLI graph construction of 57\%. Frontal-central regions (Fz, Cz, C3, C4) are identified as dominant biomarkers with Beta contribution of 58.9\% and Hjorth of 31.2\%, while Cz--T7 connectivity is consistent as a trait-level biomarker for objective screening.
\end{abstract}


\section{Introduction}

Pornography addiction in adolescents is a public health problem that is increasingly concerning in the digital era. Data from the National Survey on Children and Adolescent Life Experiences (SNPHAR) from the Ministry of Women Empowerment and Child Protection \citep{kemenpppa2021} shows that 66.6\% of boys and 62.3\% of girls aged 13--17 years in Indonesia are exposed to online pornographic content. According to \citet{faisal2022} and \citet{adarsh2023}, this condition risks disrupting prefrontal cortex development which plays a role in executive functions, decision-making, and impulse control. Therefore, implementation of protection through objective and accurate early detection methods is needed to identify adolescents who are at risk or have experienced pornography addiction so that intervention can be done as early as possible.

Conventional screening methods such as Young's Pornography Addiction Screening Test (YPAST) have significant limitations in the form of self-report bias and social desirability. Based on \citet{privara2023}, pornography consumption can induce cognitive-affective distress, guilt, and moral conflict. Therefore, objective biomarkers that can measure neurobiological changes due to pornography addiction without relying on subjective reports are needed.

\citet{huang2025} and \citet{bouhadja2022} proved that Electroencephalogram (EEG) is effective in detecting brain activity changes related to various types of addiction. This can include utilizing public Electroencephalogram datasets from adolescents with and without pornography addiction provided by \citet{kang2021}. \citet{sun2022} have shown significant advances in applying deep learning based on Convolutional Neural Network (CNN) for EEG data analysis on addiction problems. However, \citet{huang2025} noted that most other research is still dominated by traditional machine learning such as Support Vector Machine (SVM), Random Forest (RF), and Multi-Layer Perceptron (MLP) using static features, so they have not been able to capture temporal dynamics and functional connectivity between brain regions. In fact, \citet{brand2016} emphasized that the brain is a complex dynamic system with fluctuating connectivity, especially when exposed to addictive stimuli. This is reinforced by the Interaction of Person-Affect-Cognition-Execution (I-PACE) model from \citet{brand2019} which highlights the importance of understanding temporal dynamics in addictive behavior.

Graph Neural Network (GNN) has proven superior in modeling relationally structured data such as EEG signals representing brain connectivity networks \citep{abadal2025, grana2023, xue2024}. Recent reviews show GNN potential in EEG analysis for various applications, especially emotion detection \citep{almohammadi2024, klepl2024} and psychiatric disorders \citep{liu2025}, although its application for pornography addiction detection has never been explored. Therefore, this study proposes a Dynamic Spatio-Temporal Graph Neural Network (DST-GNN) architecture that integrates Graph Attention Network (GAT) for modeling spatial brain connectivity based on Phase Lag Index (PLI). According to \citet{stam2007}, this metric is chosen because it is robust against volume conduction artifacts. In addition, Bidirectional Gated Recurrent Unit (BiGRU) is also used to capture temporal dynamics of neural activity. Bidirectional RNN architecture has proven effective for EEG sequence analysis, both using Bi-LSTM \citep{algarni2022, jusseaume2022} and GRU \citep{cho2014}. \citet{walther2023} comprehensively compared various deep learning architectures for EEG time series analysis. The main novelty of this research is the first application of GNN on pornography addiction detection, integration of PLI metrics in graph construction, and dynamic modeling of brain connectivity across time.

The objectives of this study are: (1) to develop a DST-GNN model for EEG-based pornography addiction classification, (2) to identify neurobiological biomarkers in the form of brain regions and connectivity patterns that are most discriminative through explainability analysis, and (3) to analyze brain connectivity differences across various cognitive and emotional conditions.

\section{Methods}

This study uses a quantitative experimental approach with a cross-sectional design to develop and evaluate the DST-GNN model in detecting pornography addiction based on EEG signals. The research methodology is systematically arranged through several integrated stages, including: data acquisition, preprocessing, feature extraction, graph construction, model development, hyperparameter optimization, evaluation, and explainability analysis.

\begin{figure}[!htbp]
\centering
\includegraphics[width=0.95\linewidth]{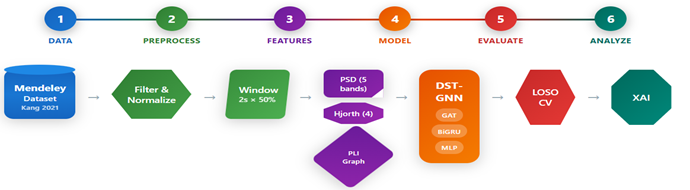}
\vspace{-2mm}
\caption{Research design flowchart}
\label{fig:pipeline}
\end{figure}

Figure~\ref{fig:pipeline} presents the research design flowchart comprehensively. The research pipeline consists of six main phases: (1) Data phase includes EEG dataset acquisition from Mendeley Data; (2) Preprocessing phase includes bandpass filter, notch, normalization, and windowing; (3) Features phase includes Power Spectral Density (PSD) extraction and Hjorth parameters as well as PLI graph construction; (4) Modeling phase includes DST-GNN and baseline model training; (5) Evaluation phase uses LOSO-CV and ablation study; (6) Analysis phase includes explainability and biomarker identification. The data flow follows transformation from 14 subjects with 19 EEG channels to produce binary classification with F1-Score 71\% and recall 86\%.

\begin{figure}[!htbp]
\centering
\includegraphics[width=0.95\linewidth]{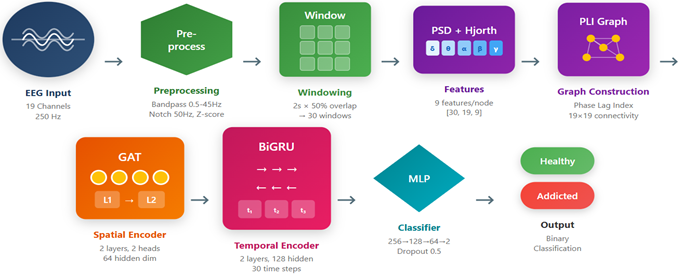}
\vspace{-2mm}
\caption{Dynamic Spatio-Temporal Graph Neural Network (DST-GNN) Architecture}
\label{fig:architecture}
\end{figure}

\FloatBarrier
Figure~\ref{fig:architecture} displays the DST-GNN architecture in detail. The model consists of three main integrated components: (1) Spatial Encoder using two Graph Attention Network (GAT) layers with 2 attention heads and hidden dimension 64 for modeling spatial connectivity between EEG channels \citep{velickovic2018, venu2024}; (2) Temporal Encoder using Bidirectional GRU (BiGRU) 2 layers with hidden dimension 128 for capturing temporal dynamics of graph sequences; and (3) Classifier in the form of Multi-Layer Perceptron (MLP) with dropout 0.5 for binary classification. Input is 19-channel EEG signals transformed into 30 temporal windows with 9 features per node (consisting of 5 PSD bands + 4 Hjorth parameters).

Spatial Encoder uses Graph Attention Network (GAT) 2-layer with multi-head attention mechanism to compute node representations based on neighbor features with adaptive attention weights according to Equations~(\ref{eq:attention}) and~(\ref{eq:node_update}):
\begin{equation}
\alpha_{ij} = \text{softmax}(\text{LeakyReLU}(\mathbf{a}^T[\mathbf{W}\mathbf{h}_i \| \mathbf{W}\mathbf{h}_j]))
\label{eq:attention}
\end{equation}
\begin{equation}
\mathbf{h}'_i = \sigma\left(\sum_{j \in \mathcal{N}_i} \alpha_{ij} \mathbf{W}\mathbf{h}_j\right)
\label{eq:node_update}
\end{equation}
where $\alpha_{ij}$ is the attention coefficient, $\mathbf{W}$ is the learnable weight matrix, $\mathbf{h}_i$ and $\mathbf{h}_j$ are node features of node $i$ and $j$, and $\mathcal{N}_i$ is the neighborhood of node $i$ in the PLI graph.

Temporal Encoder uses Bidirectional Gated Recurrent Unit (BiGRU) 2-layer to model graph sequence dynamics according to Equations~(\ref{eq:update_gate})--(\ref{eq:hidden_output}):
\begin{equation}
\mathbf{z}_t = \sigma(\mathbf{W}_z \cdot [\mathbf{h}_{t-1}, \mathbf{x}_t])
\label{eq:update_gate}
\end{equation}
\begin{equation}
\mathbf{r}_t = \sigma(\mathbf{W}_r \cdot [\mathbf{h}_{t-1}, \mathbf{x}_t])
\label{eq:reset_gate}
\end{equation}
\begin{equation}
\tilde{\mathbf{h}}_t = \tanh(\mathbf{W} \cdot [\mathbf{r}_t \odot \mathbf{h}_{t-1}, \mathbf{x}_t])
\label{eq:candidate}
\end{equation}
\begin{equation}
\mathbf{h}_t = (1 - \mathbf{z}_t) \odot \mathbf{h}_{t-1} + \mathbf{z}_t \odot \tilde{\mathbf{h}}_t
\label{eq:hidden_output}
\end{equation}
where $\mathbf{z}_t$ is the update gate, $\mathbf{r}_t$ is the reset gate, $\mathbf{h}_{t-1}$ is the previous hidden state, $\mathbf{x}_t$ is input at time $t$, $\tilde{\mathbf{h}}_t$ is the candidate hidden state, and $\mathbf{h}_t$ is the output hidden state.

Model training was performed using NVIDIA A100-SXM4-40GB GPU with 42.4 GB memory. Hyperparameter optimization uses the Optuna framework with 30 trials and automatic pruning. The best parameters were found at trial 12 with composite score (0.6$\times$accuracy + 0.4$\times$F1) = 0.6524: learning rate: 0.000668, hidden dimension: 64, GAT heads: 2, GRU layers: 2, GRU hidden: 128, dropout: 0.182, weight decay: $3.53\times10^{-5}$. AdamW optimizer with Cosine Annealing learning rate scheduler was applied for more stable convergence. Early stopping with patience 15 epochs was applied to prevent overfitting. Evaluation uses Leave-One-Subject-Out Cross-Validation (LOSO-CV) \citep{allgaier2024} with 3 random seeds (42, 123, 456) to ensure result reliability.

In addition to the DST-GNN model, five baseline models were also evaluated for comparison: Logistic Regression with L2 regularization, Support Vector Machine (SVM) with RBF (Radial Basis Function) kernel, Random Forest (RF) with 100 estimators, XGBoost with default parameters, and Multi-Layer Perceptron (MLP) with 256-128-64 neuron architecture. All baselines use the same features (PSD + Hjorth + correlation) with LOSO-CV evaluation.

\begin{figure}[!htbp]
\centering
\includegraphics[width=0.32\linewidth]{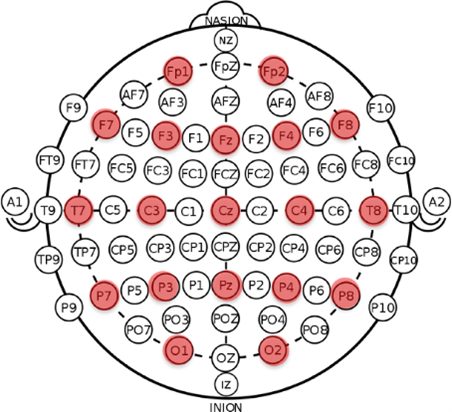}
\vspace{-2mm}
\caption{Electrode placement on scalp marked in red color (Source: \citet{kang2021})}
\label{fig:electrodes}
\end{figure}

The research dataset was obtained from Mendeley Data collected by \citet{kang2021}. The dataset consists of EEG recordings from 14 adolescent participants aged 13-15 years, divided into 7 participants with pornography addiction (YPAST score $\geq$ 36) and 7 healthy participants as controls. Recording was performed using a 19-channel EEG system with electrode placement following the International 10-20 System standard at 250 Hz sampling frequency. Electrode placement follows the International 10-20 System standard which is an international standard system for EEG electrode placement. This system divides the head into areas based on percentage of distance between anatomical reference points. Figure~\ref{fig:electrodes} shows the positions of 19 electrodes (Fp1, Fp2, F7, F3, Fz, F4, F8, T7, C3, Cz, C4, T8, P7, P3, Pz, P4, P8, O1, O2) used in EEG data acquisition.

Table~\ref{tab:eeg_details} shows details of EEG signals recorded for each subject, including task name, file name, recording duration, and data dimensions in the form of samples $\times$ channels. Each participant underwent 9 experimental conditions designed to evaluate brain responses to various stimuli: Eyes Closed (EC), Eyes Open (EO), Happy (H), Calm (C), Sad (S), Fear (F), Memorize (M), Executive Task (ET) with pornographic stimulus exposure, and Recall (R). Total recording duration per subject is 10 minutes with 250 Hz sampling frequency.

\begin{table}[!htbp]
\caption{EEG signal recording details per experimental condition (Source: \citet{kang2021})}
\label{tab:eeg_details}
\centering
\small
\begin{tabular}{@{}clccc@{}}
\toprule
No. & Task Name & File Name & Duration (s) & Samples $\times$ Channels \\
\midrule
1 & Eyes Closed & EC.csv & 60 & 15,000 $\times$ 19 \\
2 & Eyes Open & EO.csv & 60 & 15,000 $\times$ 19 \\
3 & Happy & H.csv & 60 & 15,000 $\times$ 19 \\
4 & Calm & C.csv & 60 & 15,000 $\times$ 19 \\
5 & Sad & S.csv & 60 & 15,000 $\times$ 19 \\
6 & Fear & F.csv & 60 & 15,000 $\times$ 19 \\
7 & Memorise 15 words & M.csv & 60 & 15,000 $\times$ 19 \\
8 & Executive Tasks & ET.csv & 120 & 30,000 $\times$ 19 \\
9 & Recall 15 words & R.csv & 60 & 15,000 $\times$ 19 \\
\bottomrule
\end{tabular}
\end{table}

\begin{figure}[!htbp]
\centering
\includegraphics[width=0.38\linewidth]{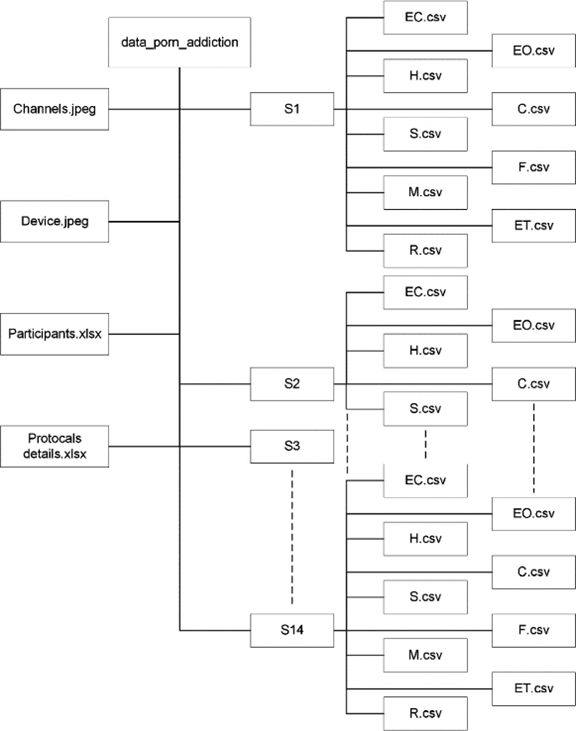}
\vspace{-2mm}
\caption{Dataset folder structure (Source: \citet{kang2021})}
\label{fig:folder}
\end{figure}

The dataset is stored in an organized folder structure with the main folder data\_porn\_addiction containing 14 subfolders (S1-S14) for each subject as shown in Figure~\ref{fig:folder}. Each subject folder contains .csv files for each experimental task (EC.csv, EO.csv, H.csv, C.csv, S.csv, F.csv, M.csv, ET.csv, R.csv). Additional information such as channel representation, acquisition device, participant labels, and experimental protocol details are also available.

Table~\ref{tab:participants} shows participant details involved in data collection, including subject ID, gender, and classification label (Addicted or Not Addicted) based on YPAST screening results verified by clinical psychologists.

\begin{table}[!htbp]
\caption{Participant details (Source: \citet{kang2021})}
\label{tab:participants}
\centering
\small
\begin{tabular}{@{}clll@{}}
\toprule
No. & Subject ID & Gender & Label \\
\midrule
1 & S1 & Male & Addicted \\
2 & S2 & Female & Not Addicted \\
3 & S3 & Female & Not Addicted \\
4 & S4 & Male & Not Addicted \\
5 & S5 & Male & Addicted \\
6 & S6 & Male & Addicted \\
7 & S7 & Male & Not Addicted \\
8 & S8 & Male & Not Addicted \\
9 & S9 & Female & Addicted \\
10 & S10 & Female & Addicted \\
11 & S11 & Female & Addicted \\
12 & S12 & Male & Not Addicted \\
13 & S13 & Male & Not Addicted \\
14 & S14 & Male & Addicted \\
\bottomrule
\end{tabular}
\end{table}

\begin{figure}[!htbp]
\centering
\includegraphics[width=0.65\linewidth]{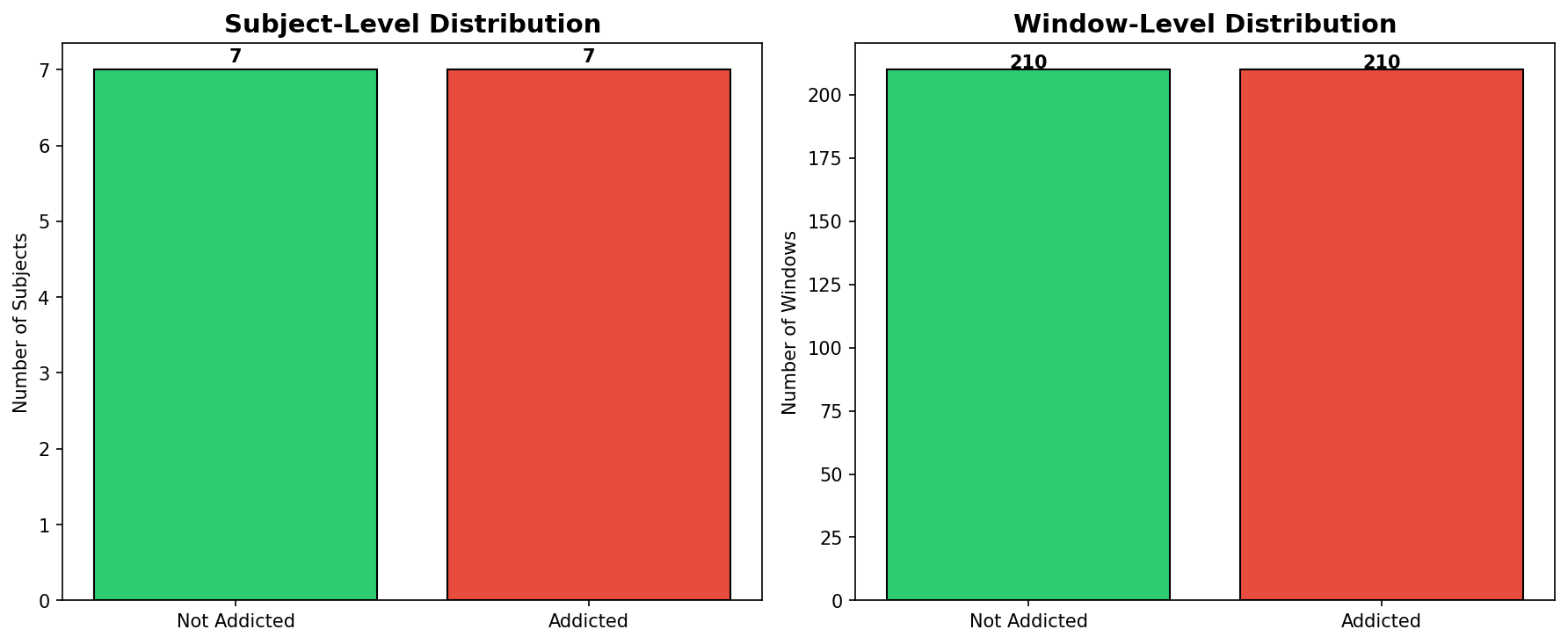}
\vspace{-2mm}
\caption{Data distribution at subject level (left) and window level (right)}
\label{fig:distribution}
\end{figure}

The ET condition is the main focus of analysis because it involves direct exposure to pornographic stimuli, but multi-condition analysis was also performed to identify consistent biomarkers. Figure~\ref{fig:distribution} shows balanced data distribution at the subject level (7:7) with a total of 420 temporal windows after preprocessing.

\begin{figure}[!htbp]
\centering
\includegraphics[width=0.78\linewidth]{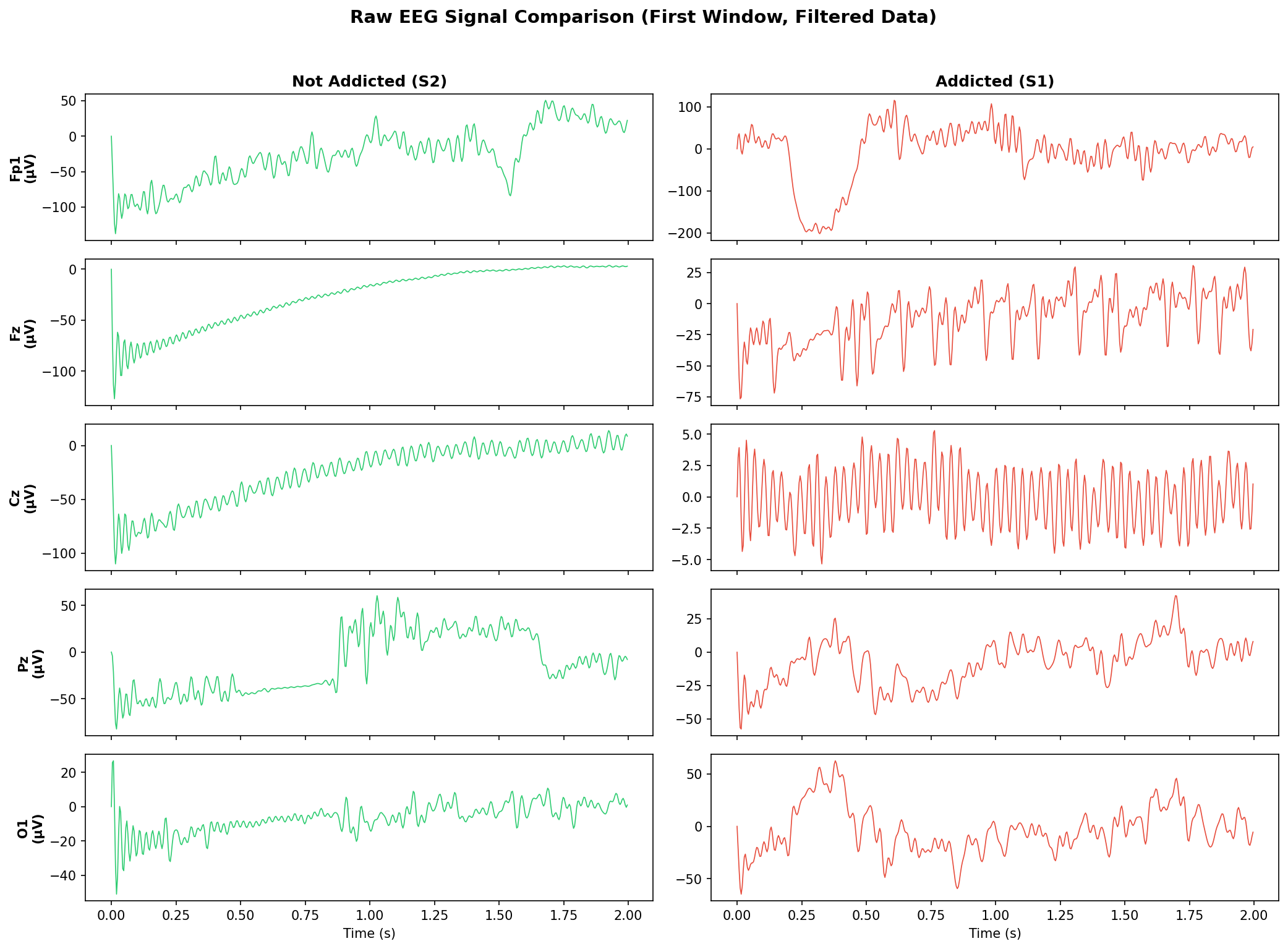}
\vspace{-2mm}
\caption{Comparison of filtered EEG signals between healthy and addicted subjects}
\label{fig:eeg_comparison}
\end{figure}

EEG signal preprocessing was performed through several stages. First, a 4th order Butterworth bandpass filter was applied with cutoff frequency 0.5-45 Hz to remove low frequency noise (drift) and high frequency noise (muscle artifact, powerline). A notch filter at 50 Hz was applied to remove electrical interference. Furthermore, signals were normalized using z-score normalization per channel calculated according to Equation~(\ref{eq:zscore}):
\begin{equation}
z = \frac{x - \mu}{\sigma}
\label{eq:zscore}
\end{equation}
where $x$ is the amplitude value, $\mu$ is the mean, and $\sigma$ is the channel standard deviation. Figure~\ref{fig:eeg_comparison} shows comparison of filtered EEG signals (first window, 2 seconds) between healthy and addicted subjects on five representative channels (Fp1, Fz, Cz, Pz, O1), with the left panel showing non-addicted subject (S2) in green and the right panel addicted subject (S1) in orange-red, y-axis showing amplitude ($\mu$V) and x-axis time (seconds), which generally shows that the addicted group has higher and more irregular amplitude fluctuations---especially on Fp1 channel---while the healthy group displays more stable drift patterns on Fz and more consistent Cz oscillations compared to high variability in the addicted group.

Two categories of features were extracted from each EEG window. First, PSD was calculated using Welch's method \citep{chaddad2023, welch1967} on 5 standard frequency bands: Delta (0.5--4 Hz), Theta (4--8 Hz), Alpha (8--13 Hz), Beta (13--30 Hz), and Gamma (30--45 Hz). Second, Hjorth parameters \citep{hjorth1970} were extracted including activity (signal variance), mobility (ratio of first derivative standard deviation to original signal), complexity (ratio of first derivative mobility to signal mobility), and mean amplitude. A total of 9 features were extracted per channel, yielding 171 node features.

\begin{figure}[!htbp]
\centering
\includegraphics[width=0.80\linewidth]{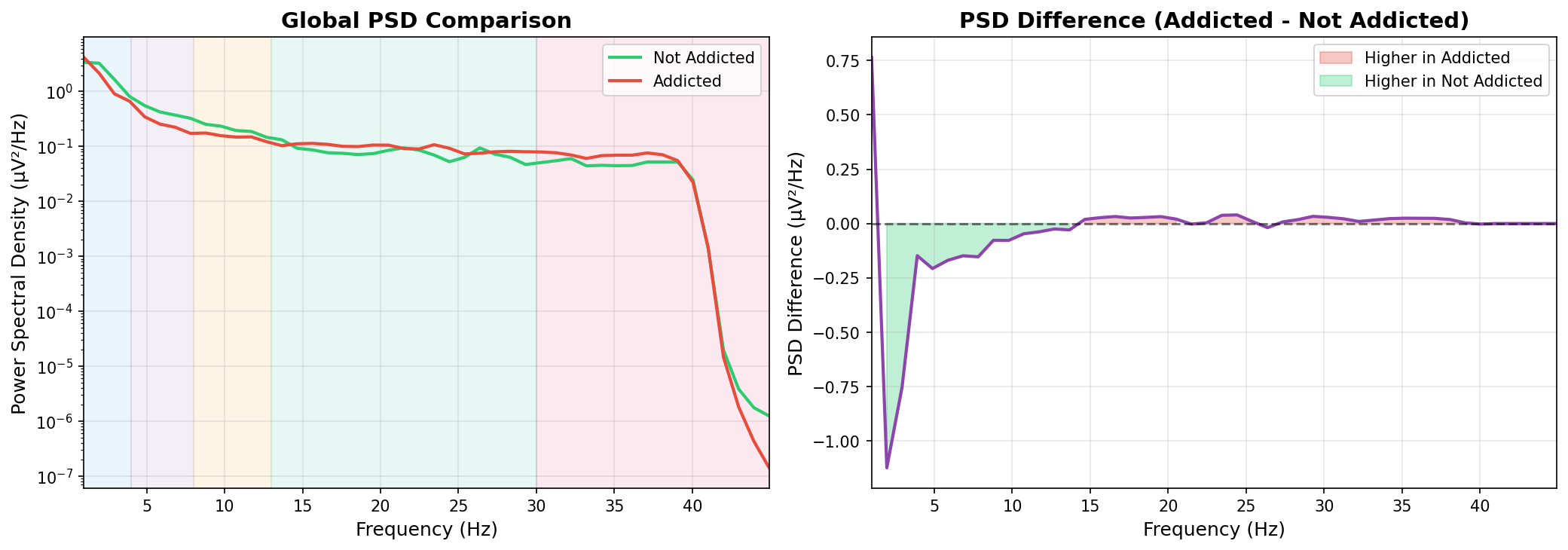}
\vspace{-2mm}
\caption{Global PSD (Power Spectral Density) comparison between groups}
\label{fig:psd}
\end{figure}

\begin{figure}[!htbp]
\centering
\includegraphics[width=0.62\linewidth]{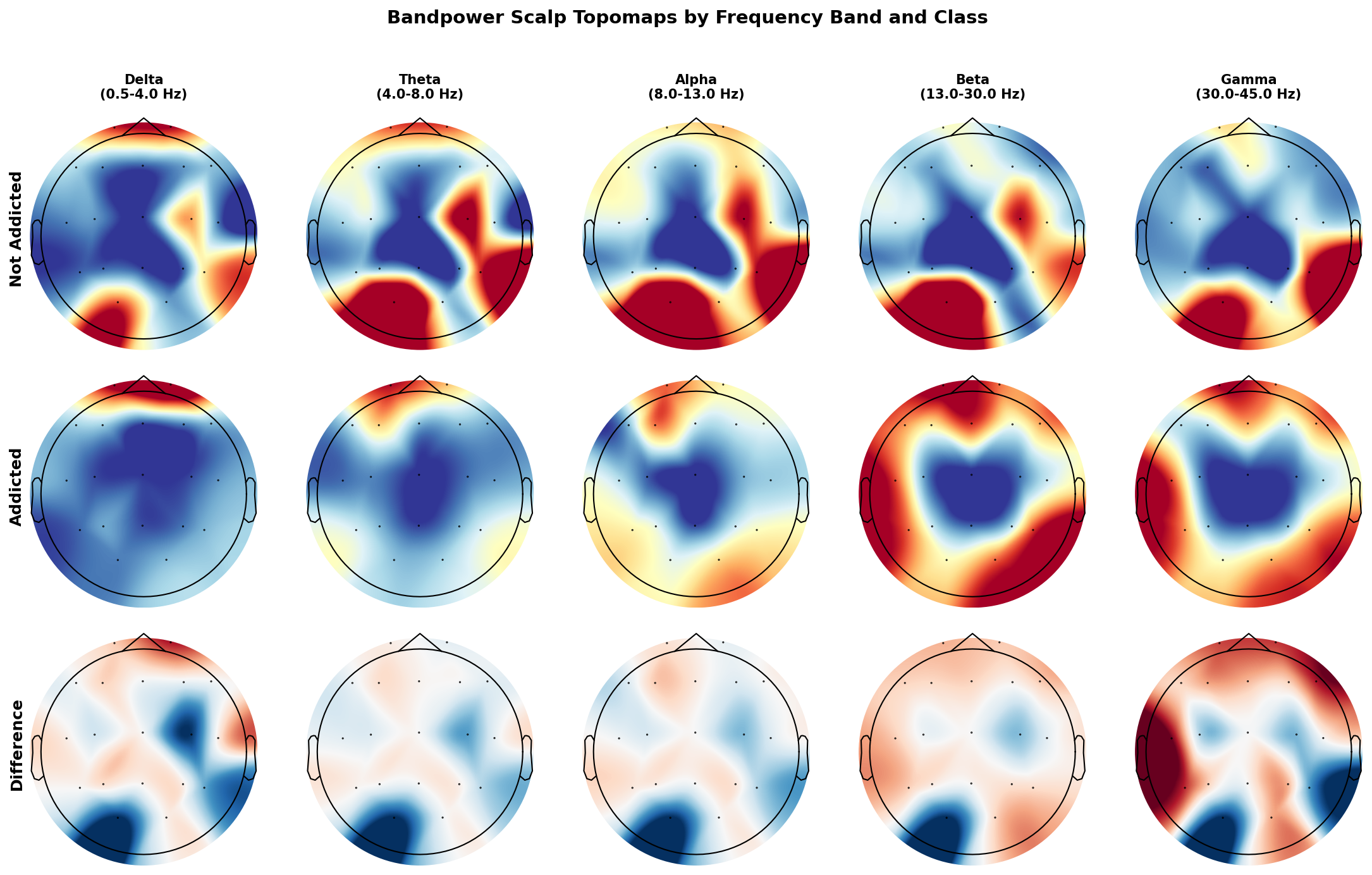}
\vspace{-2mm}
\caption{Scalp topomaps band power per frequency: healthy (top), addicted (middle), difference (bottom)}
\label{fig:topomaps}
\end{figure}

Figures~\ref{fig:psd} and~\ref{fig:topomaps} present comprehensive spectral analysis revealing important differences between the two groups. Figure~\ref{fig:psd} consists of two panels: the left panel shows global PSD comparison in logarithmic scale with green line (not addicted) and red (addicted) above colored background showing frequency bands (Light Blue: Delta (~0 - 4 Hz), Light Purple: Theta (~4 - 8 Hz), Light Orange: Alpha (~8 - 13 Hz), Light Green: Beta (~13 - 30 Hz), Pink: Gamma (> 30 Hz)); the right panel shows PSD difference with red shading (higher in addicted) and green (higher in not addicted). It can be seen that the addicted group shows increased power in the Beta band (13-30 Hz) associated with arousal and cognitive processes, consistent with prefrontal cortex hyperactivation theory in addiction. Figure~\ref{fig:topomaps} displays a 3$\times$5 matrix of scalp topomaps showing spatial distribution of band power, with the most significant differences concentrated in frontal and central regions.

\FloatBarrier

Brain connectivity graphs were constructed using Phase Lag Index (PLI) which measures the asymmetry of phase difference distribution between two signals \citep{stam2007}. PLI is defined according to Equation~(\ref{eq:pli}):
\begin{equation}
\text{PLI} = \left| \langle \text{sign}(\Delta\phi(t)) \rangle \right|
\label{eq:pli}
\end{equation}
where $\Delta\phi(t)$ is the instantaneous phase difference between two signals at time $t$, and $\langle\cdot\rangle$ indicates temporal average. PLI ranges from 0-1, with high values indicating strong functional connectivity \citep{stam2007}. The advantage of PLI over conventional coherence metrics is its robustness against volume conduction artifacts \citep{krukow2024, polat2023}. The application of PLI in adolescent neuroimaging contexts has also been validated by recent research \citep{raveendran2025}. Additionally, weighted PLI (wPLI) was also calculated according to Equation~(\ref{eq:wpli}) for more robust multi-condition analysis \citep{yan2021}:
\begin{equation}
\text{wPLI} = \frac{\left| \langle|\Delta\phi(t)| \cdot \text{sign}(\Delta\phi(t))\rangle \right|}{\langle|\Delta\phi(t)|\rangle}
\label{eq:wpli}
\end{equation}

\begin{figure}[!htbp]
\centering
\includegraphics[width=0.85\linewidth]{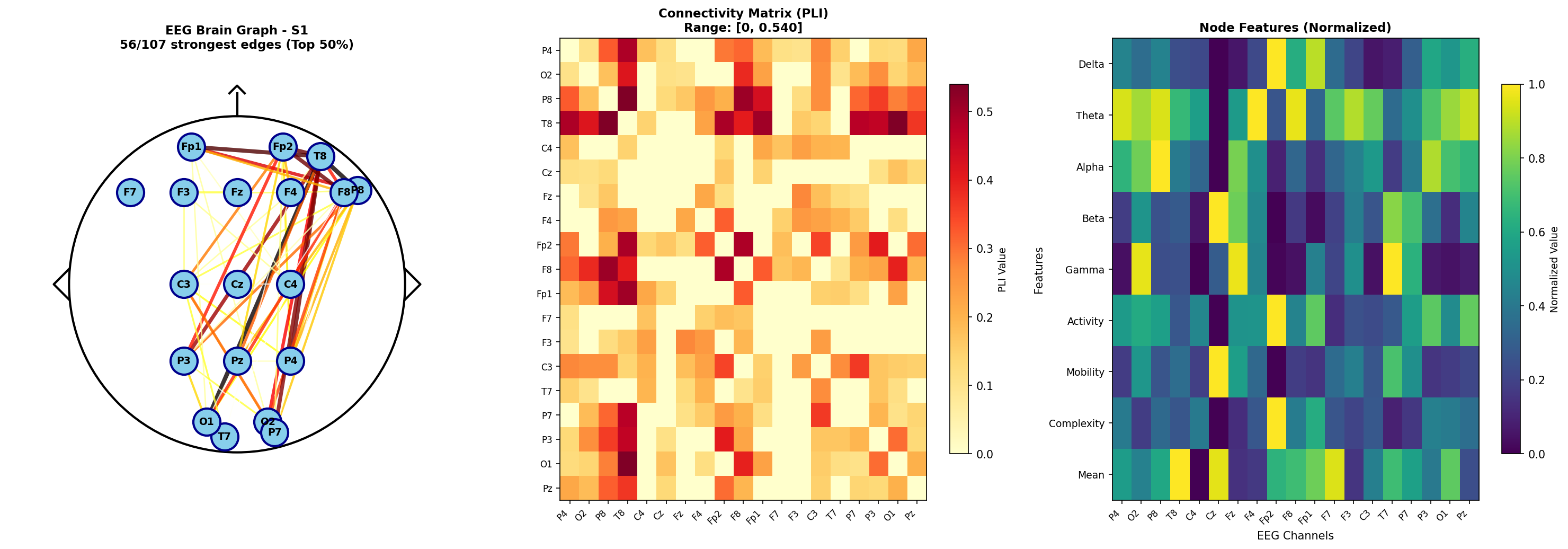}
\vspace{-2mm}
\caption{EEG graph structure: connection topology (left), PLI matrix (center), node features (right)}
\label{fig:graph_structure}
\end{figure}

\begin{figure}[!htbp]
\centering
\includegraphics[width=0.92\linewidth]{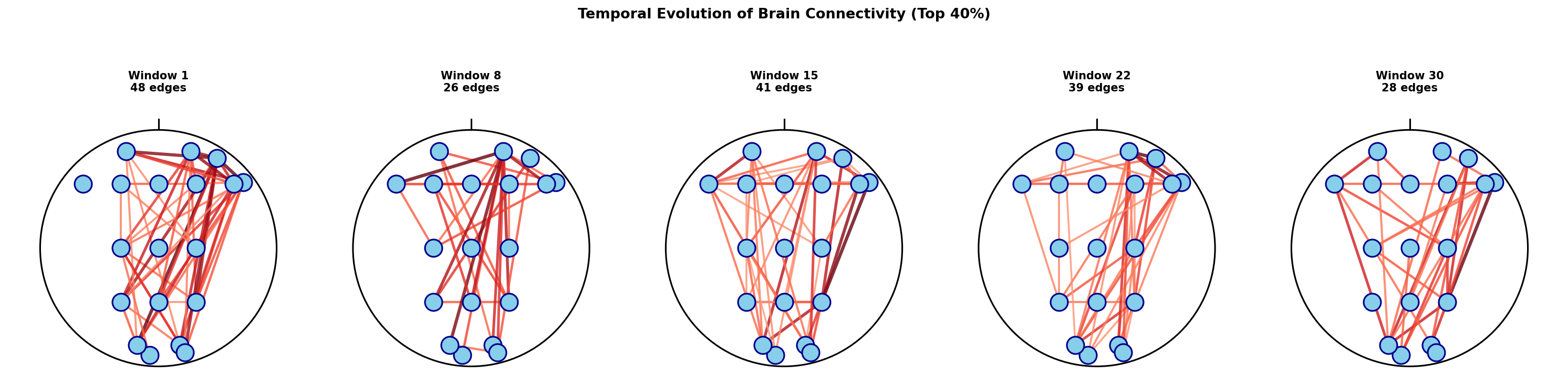}
\vspace{-2mm}
\caption{Temporal evolution of brain connectivity on window sequence}
\label{fig:temporal_evolution}
\end{figure}

Graph $G = (V, E)$ was constructed with $V$ as the set of 19 nodes (EEG channels) and $E$ as PLI-weighted edges. Thresholding at the 50th percentile was applied to retain the most significant connections. Figure~\ref{fig:graph_structure} illustrates the EEG graph construction of subject 1 (addicted) through visualization of 19-channel graph topology based on Phase Lag Index (PLI) values showing 56 of 107 strongest edges (Top 50\%), 19$\times$19 PLI matrix representing functional connectivity strength, and node feature heatmap consisting of 9 features (5 PSD bands and 4 Hjorth parameters) on each channel, while Figure~\ref{fig:temporal_evolution} shows temporal evolution of brain connectivity of the same subject on several time windows, confirming the dynamic character of functional connectivity successfully captured by the model.

\begin{figure}[!htbp]
\centering
\includegraphics[width=0.68\linewidth]{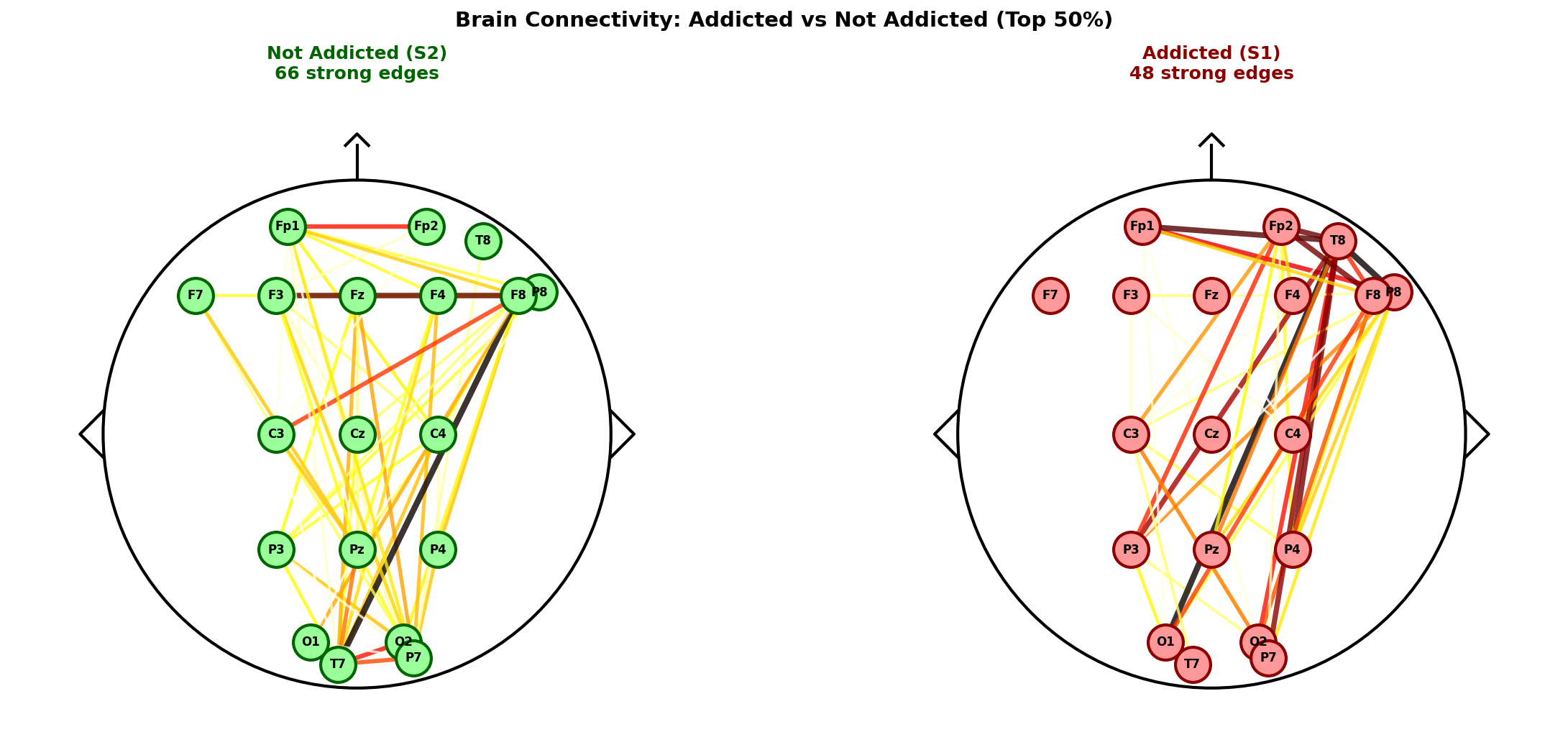}
\vspace{-2mm}
\caption{Brain connectivity pattern comparison: healthy (left, 66 edges) vs addicted (right, 48 edges)}
\label{fig:connectivity_comparison}
\end{figure}

\FloatBarrier
Figure~\ref{fig:connectivity_comparison} presents direct comparison of brain connectivity patterns between two subjects using top 50\% strongest connection graph visualization with different color schemes to distinguish groups. Analysis shows substantial differences: healthy subject (S2) has 66 strong edges with more evenly distributed connectivity patterns across brain regions, while addicted subject (S1) only has 48 strong edges with higher concentration in the right frontal-temporal region (F4, F8, T8). This pattern indicates brain network reorganization in addicted individuals, where there is decreased global connectivity but increased local clustering in regions involved in reward processing and impulse control. These findings are consistent with allostatic theory of addiction showing a shift from homeostatic regulation to pathological conditions.

\section{Results and Discussion}

This section presents comprehensive evaluation results of the DST-GNN model compared to baselines, ablation analysis to validate component contributions, and explainability analysis for model decision interpretation. Results are discussed in the context of neuroimaging findings related to addiction and implications for child protection. The complete implementation is available at the GitHub repository: \url{https://github.com/achmadardanip/eeg-research/tree/master}.

Table~\ref{tab:baseline} shows baseline model evaluation results using LOSO-CV. Logistic Regression achieves the best performance with 62.14\% accuracy and 34.73\% F1-Score. All baseline models show low recall ($<$30\%), indicating difficulty in detecting positive cases (addiction). This is due to limitations of static features that cannot capture temporal dynamics of brain connectivity.

\begin{table}[!htbp]
\caption{Baseline model evaluation results (LOSO-CV)}
\label{tab:baseline}
\centering
\small
\begin{tabular}{@{}lccccc@{}}
\toprule
Model & Accuracy & Precision & Recall & F1-Score & ROC-AUC \\
\midrule
Logistic Regression & 62.14\% & 50.00\% & 29.29\% & 34.73\% & 50.00\% \\
SVM (RBF) & 58.33\% & 50.00\% & 28.57\% & 34.37\% & 50.00\% \\
Random Forest & 44.52\% & 42.86\% & 16.43\% & 20.63\% & 50.00\% \\
XGBoost & 49.29\% & 50.00\% & 20.71\% & 26.04\% & 50.00\% \\
MLP & 51.90\% & 50.00\% & 26.19\% & 32.46\% & 50.00\% \\
\bottomrule
\end{tabular}
\end{table}

\begin{figure}[!htbp]
\centering
\includegraphics[width=0.58\linewidth]{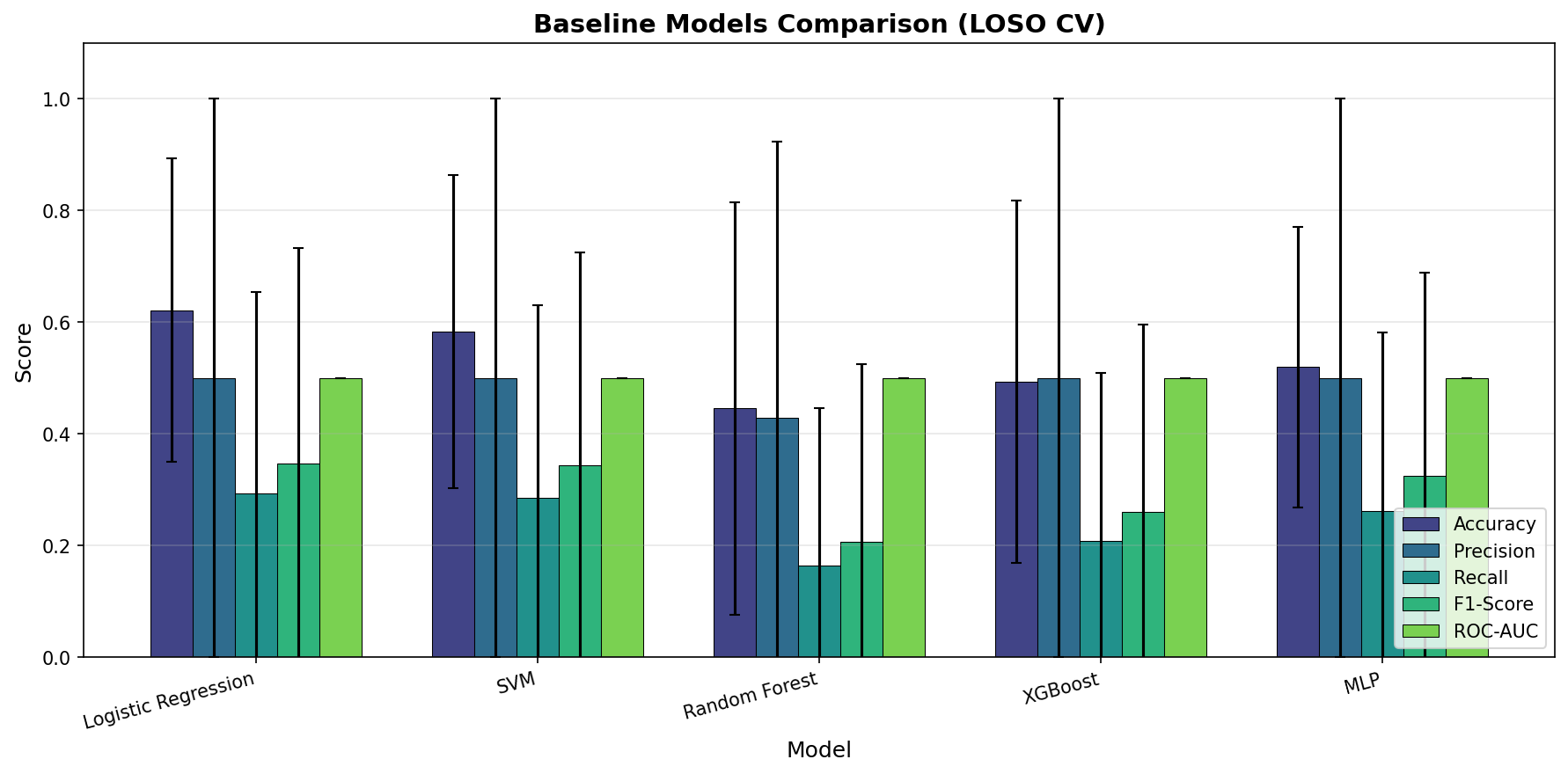}
\vspace{-2mm}
\caption{Baseline model performance comparison with error bars}
\label{fig:baseline_comparison}
\end{figure}

Figure~\ref{fig:baseline_comparison} visualizes baseline performance comparison with error bars. It can be seen that all models show high variability and suboptimal performance, especially on the recall metric which is critical for medical screening contexts where minimizing false negatives is very important.

Table~\ref{tab:dstgnn} shows DST-GNN evaluation results with multi-seed LOSO-CV. DST-GNN achieves an average F1-Score of 71.00\% $\pm$ 12.10\%, a 104\% improvement compared to the best baseline. High recall (85.71\%) is very important for screening contexts where minimizing false negatives is prioritized. Inter-seed variability reflects sensitivity to initialization, but performance is consistently better than baselines.

\begin{table}[!htbp]
\caption{DST-GNN multi-seed LOSO-CV evaluation results}
\label{tab:dstgnn}
\centering
\footnotesize
\setlength{\tabcolsep}{4pt}
\begin{tabular}{@{}lccccc@{}}
\toprule
Seed & Accuracy & Precision & Recall & F1-Score & ROC-AUC \\
\midrule
42 & 57.14\% & 54.55\% & 85.71\% & 66.67\% & 59.18\% \\
123 & 78.57\% & 77.78\% & 100.00\% & 87.50\% & 89.80\% \\
456 & 57.14\% & 50.00\% & 71.43\% & 58.82\% & 44.90\% \\
\midrule
Mean$\pm$SD & 64.29$\pm$15.43\% & 60.77$\pm$12.17\% & 85.71$\pm$11.66\% & 71.00$\pm$12.10\% & 64.63$\pm$18.73\% \\
\bottomrule
\end{tabular}
\end{table}

\begin{figure}[!htbp]
\centering
\includegraphics[width=0.78\linewidth]{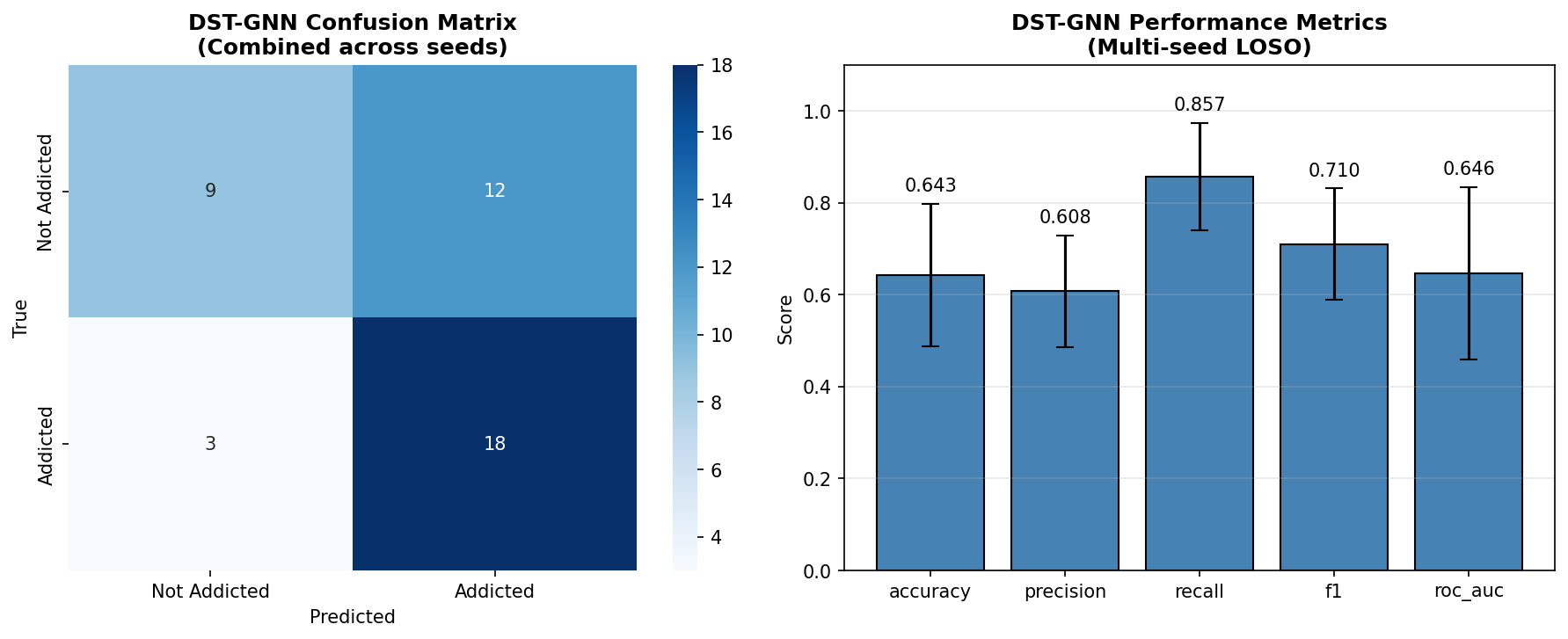}
\vspace{-2mm}
\caption{Confusion matrix (left) and DST-GNN performance metrics (right)}
\label{fig:confusion_matrix}
\end{figure}

\begin{figure}[!htbp]
\centering
\includegraphics[width=0.42\linewidth]{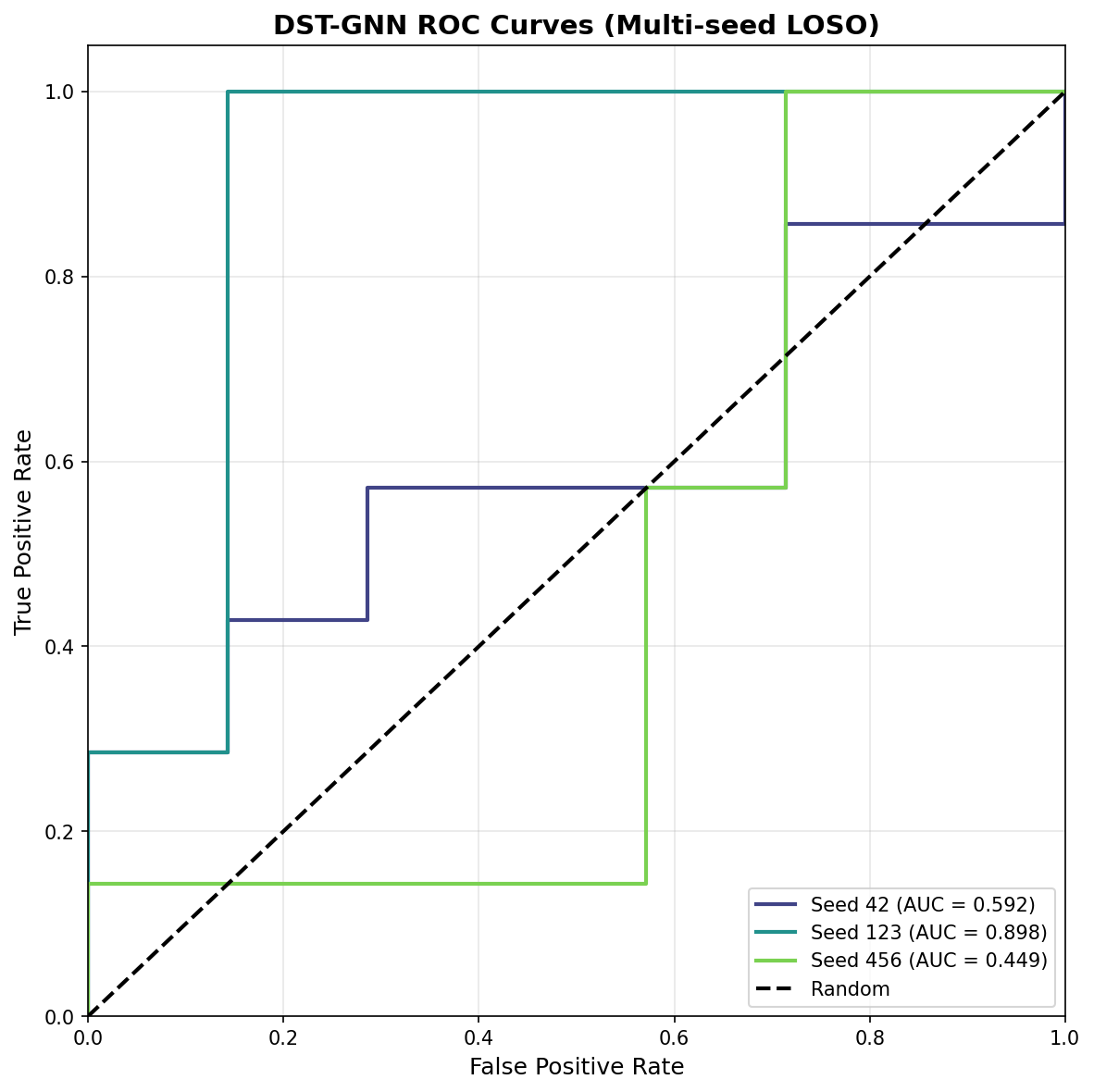}
\vspace{-2mm}
\caption{DST-GNN ROC curves for each random seed}
\label{fig:roc_curves}
\end{figure}

Figure~\ref{fig:confusion_matrix} displays the confusion matrix and DST-GNN performance metrics. Prediction distribution shows that the model tends to classify to the positive class (addiction), yielding high recall but with a trade-off on precision. Figure~\ref{fig:roc_curves} displays ROC curves for each seed, with seed 123 achieving the highest Area Under the Curve (AUC) of 0.898.

\FloatBarrier

Ablation study was conducted to validate the contribution of each DST-GNN component. Table~\ref{tab:ablation} shows that removal of temporal component (BiGRU) decreases F1-Score by 21\%, confirming the importance of modeling temporal dynamics of brain activity. Replacing PLI graph with fully-connected graph causes drastic F1-Score decrease (57\%), showing that functional connectivity-based graph construction is very critical for model performance.

\begin{table}[!htbp]
\caption{DST-GNN ablation study results}
\label{tab:ablation}
\centering
\begin{tabular}{@{}lccc@{}}
\toprule
Configuration & Accuracy & F1-Score & F1 Change \\
\midrule
Full DST-GNN & 64.29\% & 71.00\% & Baseline \\
Spatial-Only (without BiGRU) & 42.86\% & 50.00\% & $-$21.00\% \\
Fully-Connected (without PLI) & 14.29\% & 14.29\% & $-$56.71\% \\
\bottomrule
\end{tabular}
\end{table}

\begin{figure}[!htbp]
\centering
\includegraphics[width=0.58\linewidth]{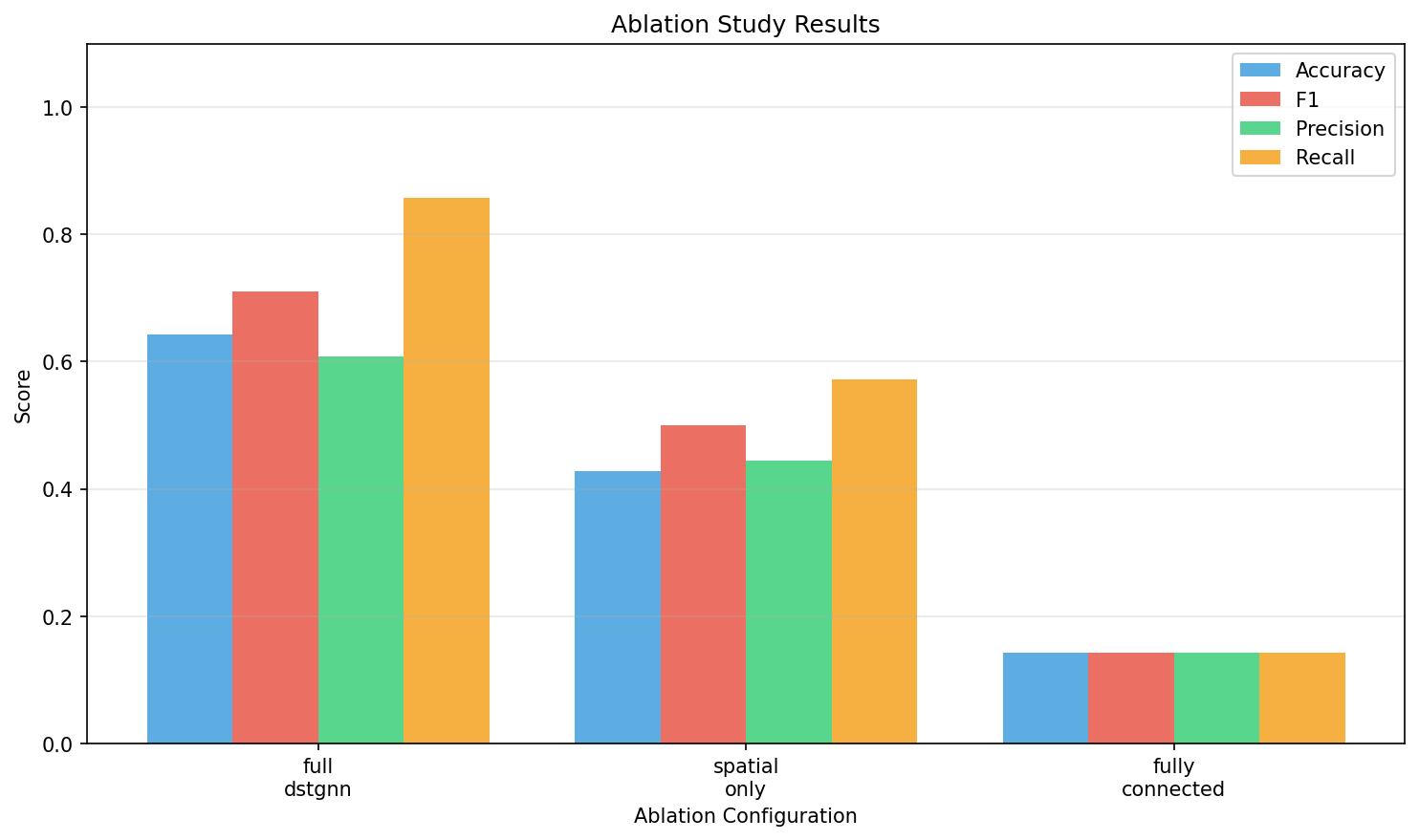}
\vspace{-2mm}
\caption{Ablation study results comparison between model configurations}
\label{fig:ablation}
\end{figure}

\FloatBarrier

\begin{figure}[!htbp]
\centering
\includegraphics[width=0.32\linewidth]{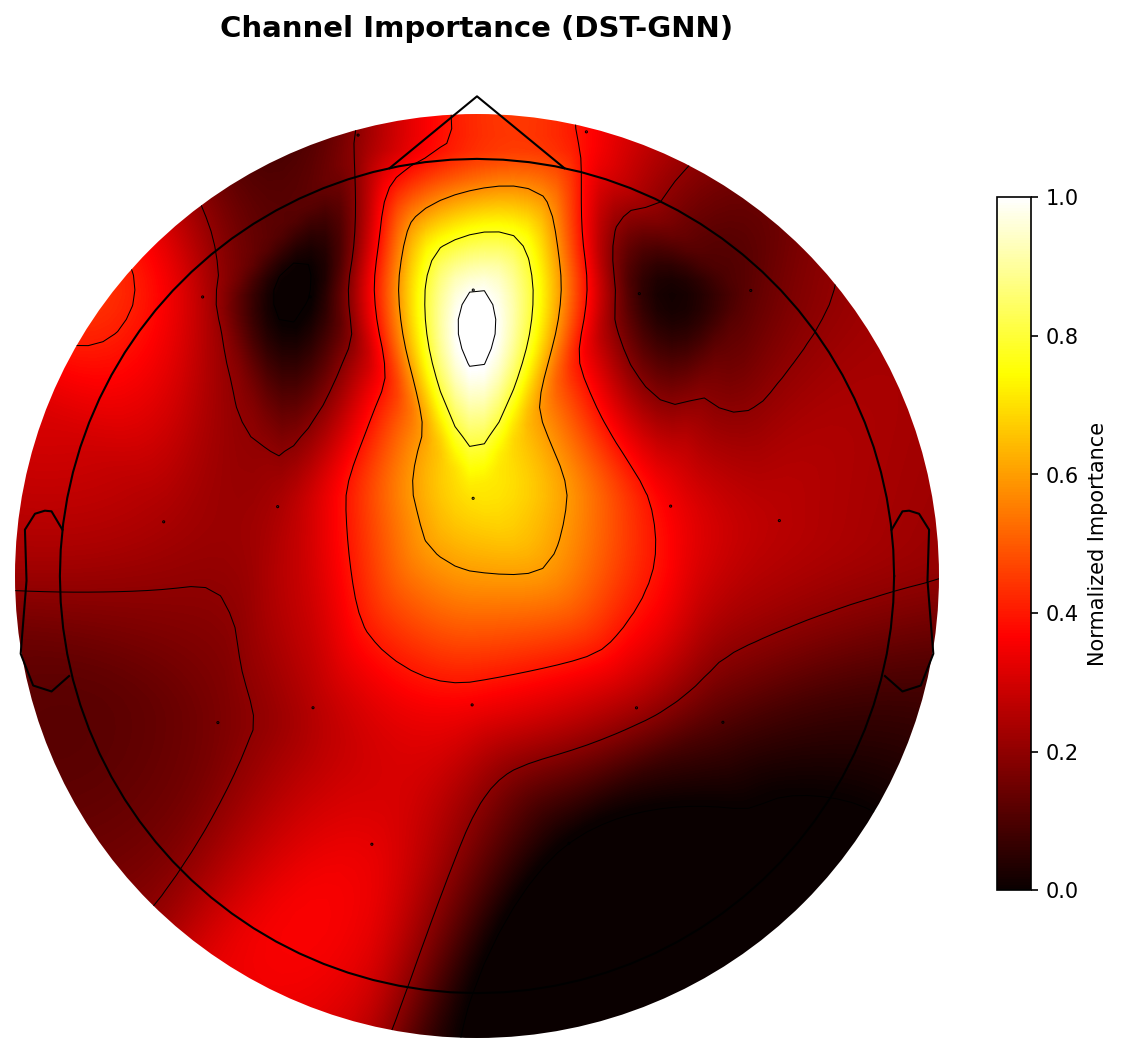}
\vspace{-2mm}
\caption{DST-GNN channel importance topomap showing frontal-central region dominance}
\label{fig:topomap_importance}
\end{figure}

Explainability analysis was conducted to identify brain regions and features that contribute most to model predictions. Figure~\ref{fig:topomap_importance} shows channel importance topomap where frontal (Fz, Fp2) and central (Cz, C3, C4) regions have the highest importance. These findings are consistent with neuroimaging literature showing prefrontal cortex involvement in impulse regulation and reward processing in addiction.

\begin{figure}[!htbp]
\centering
\includegraphics[width=0.68\linewidth]{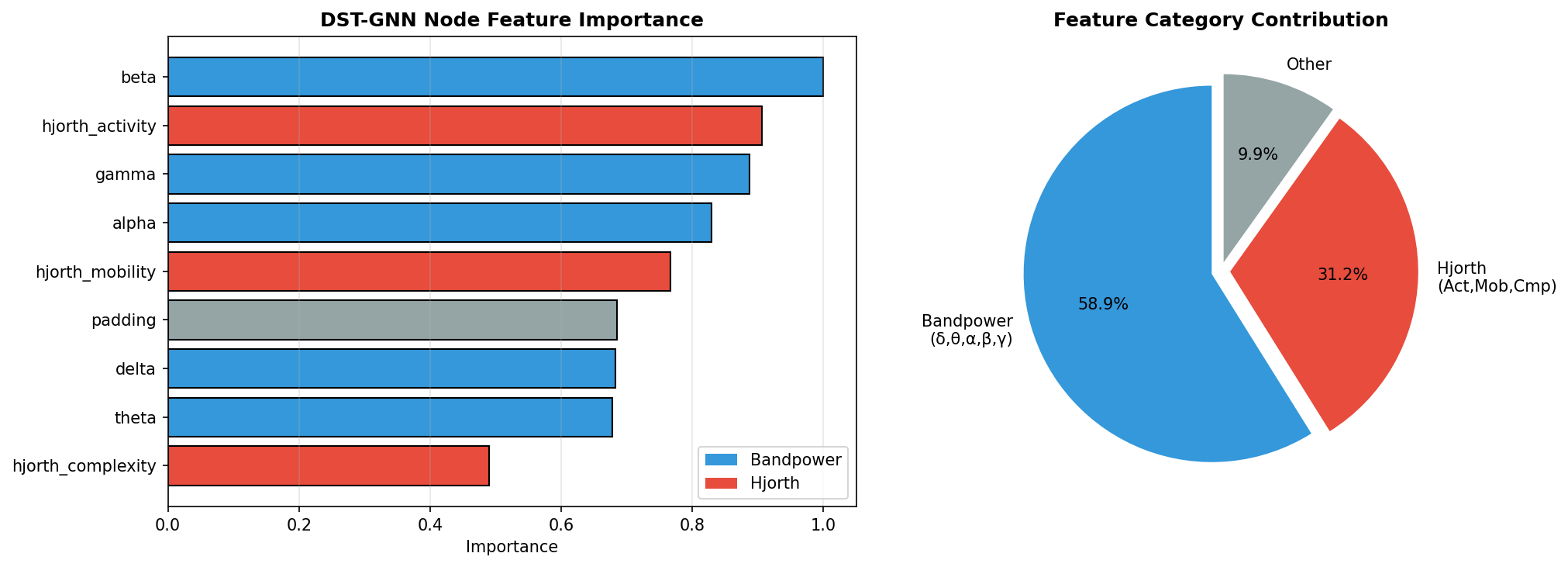}
\vspace{-2mm}
\caption{DST-GNN feature importance: Beta and Hjorth Activity as main features}
\label{fig:feature_importance}
\end{figure}

Figure~\ref{fig:feature_importance} displays feature importance revealing the largest contribution from the Beta frequency band (58.9\%) associated with cognitive processes and arousal, followed by Hjorth parameters (31.2\%) measuring temporal signal characteristics. Beta dominance is consistent with findings that addiction causes changes in high-frequency activity related to cognitive control.

\FloatBarrier

\begin{figure}[!htbp]
\centering
\includegraphics[width=0.58\linewidth]{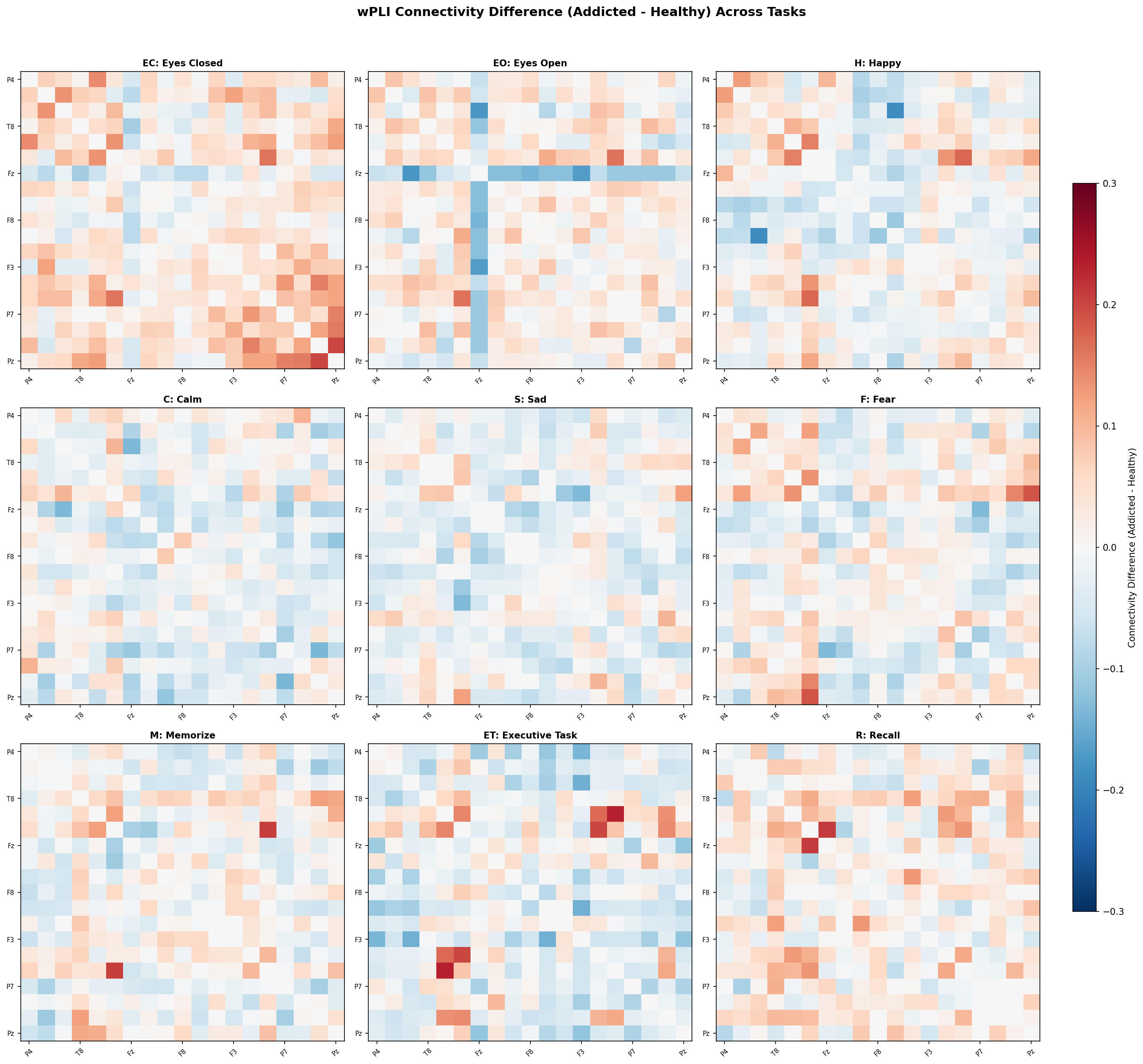}
\vspace{-2mm}
\caption{wPLI connectivity differences (Addicted - Healthy) on 9 experimental conditions}
\label{fig:wpli_diff}
\end{figure}

Functional connectivity analysis was performed on all 9 experimental conditions using wPLI to identify consistent and condition-specific patterns. Figure~\ref{fig:wpli_diff} shows wPLI connectivity differences between addicted and healthy groups on 9 experimental conditions. Red color indicates increased connectivity in the addicted group, while blue indicates decrease. In baseline conditions (EC, EO), increased global connectivity is seen in the addicted group. Emotional conditions (H, C, S, F) show heterogeneous patterns with increase in central-temporal connections. Cognitive conditions (M, R) and Executive Task (ET) show the most significant increase in C4-T7, Cz-C3, and C4-C3 connections, indicating hyperconnectivity in regions involved in reward processing and cognitive control.

An important finding of this research is the Cz-T7 connection which shows consistent differences across various conditions (EC, EO, H, M), indicating a trait-level biomarker that does not depend on specific experimental conditions. Several studies report functional brain activity changes related to pornography consumption and addiction, particularly involving temporal and frontal regions \citep{prantner2024, privara2023, shu2025}. These findings open opportunities for developing objective detection approaches that do not depend on pornographic stimulus exposure during examination.

\begin{figure}[!htbp]
\centering
\includegraphics[width=0.92\linewidth]{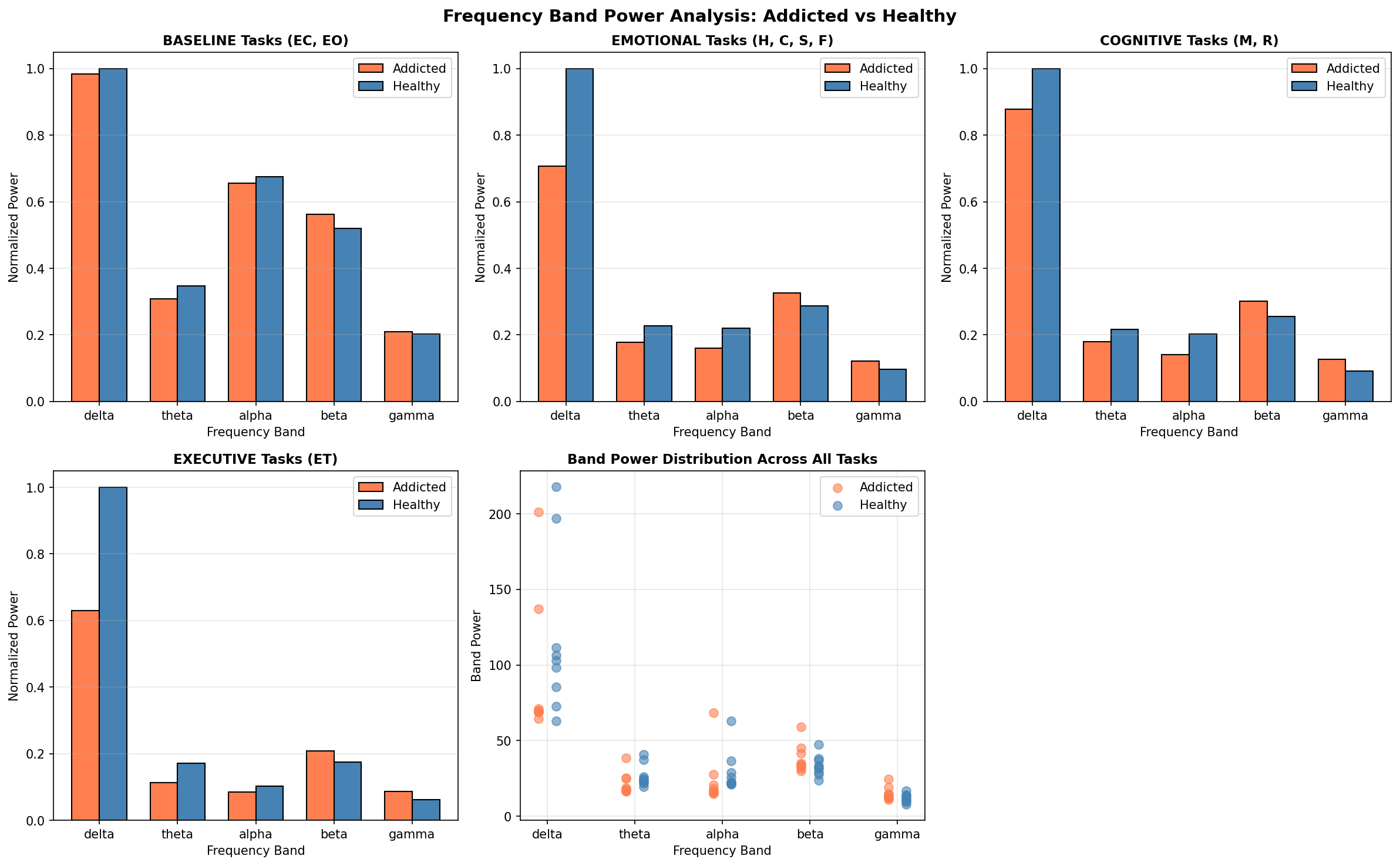}
\vspace{-2mm}
\caption{Band power analysis per condition category: baseline, emotional, cognitive, executive}
\label{fig:band_power}
\end{figure}

Figure~\ref{fig:band_power} presents band power analysis per condition category. The addicted group shows lower power in Alpha and Theta bands across various conditions. Significant differences ($p<0.05$) were found in memorize condition (Alpha $p=0.0002$; Theta $p=0.0014$), fear (Alpha $p=0.0029$; Gamma $p=0.0192$ higher in addicted), and calm (Alpha $p=0.0044$). Decreased Alpha is related to reduced emotional regulation ability, consistent with the I-PACE model.

\FloatBarrier

\begin{figure}[!htbp]
\centering
\includegraphics[width=0.92\linewidth]{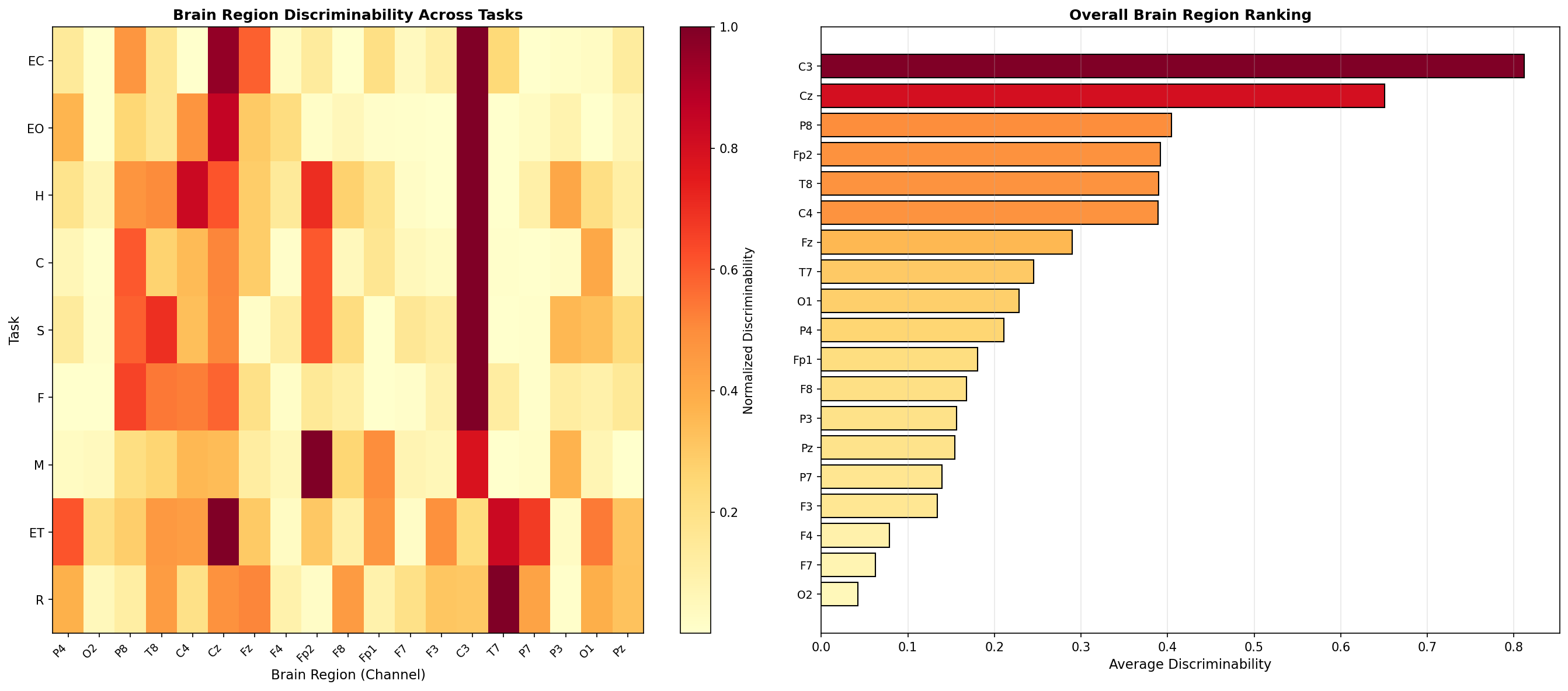}
\vspace{-2mm}
\caption{Brain region discriminability between tasks (left) and overall ranking (right)}
\label{fig:discriminability}
\end{figure}

\begin{figure}[!htbp]
\centering
\includegraphics[width=0.92\linewidth]{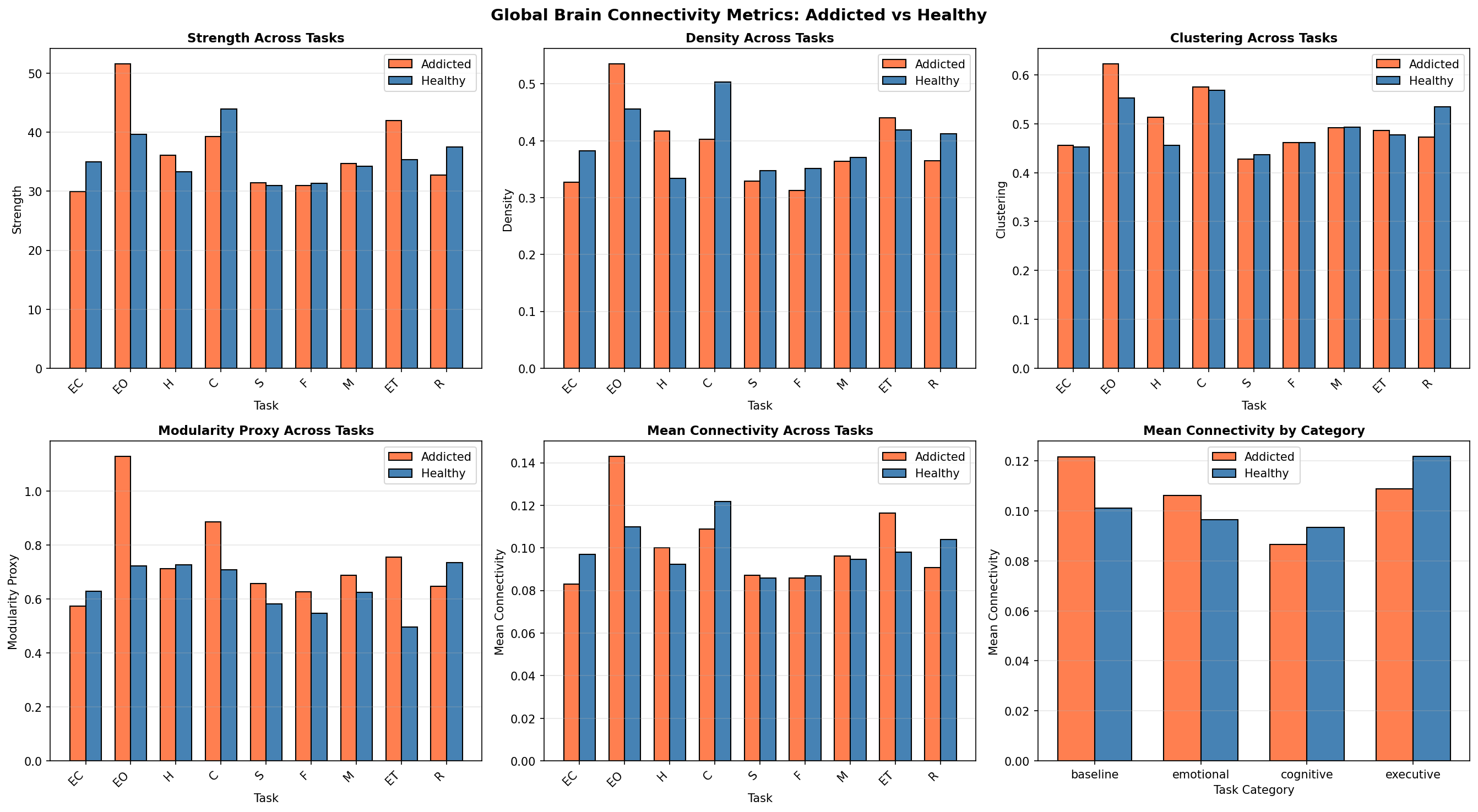}
\vspace{-2mm}
\caption{Global connectivity metrics between conditions and task categories}
\label{fig:global_connectivity}
\end{figure}

Figure~\ref{fig:discriminability} displays brain region discriminability between tasks, confirming that central regions (C3, Cz, C4) are most consistently discriminative across various conditions. Figure~\ref{fig:global_connectivity} shows global connectivity metrics revealing interesting patterns: the addicted group has higher connectivity in baseline conditions but lower in emotional conditions, indicating context-dependent network dysregulation.

\FloatBarrier

\begin{figure}[!htbp]
\centering
\includegraphics[width=0.92\linewidth]{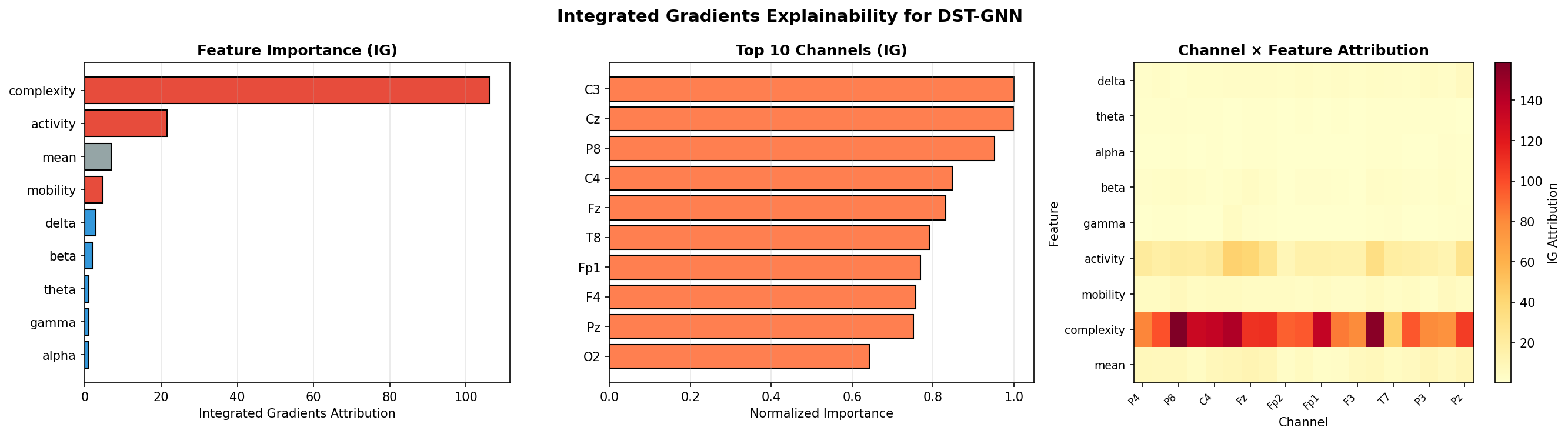}
\vspace{-2mm}
\caption{Integrated Gradients (IG): feature importance (left), top channels (center), channel$\times$feature (right)}
\label{fig:integrated_gradients}
\end{figure}

Advanced explainability analysis using Integrated Gradients (IG) \citep{cui2023, farahani2022} and gradient-based edge importance provides more detailed insights. Figure~\ref{fig:integrated_gradients} confirms Hjorth complexity as the most important feature, with top channels C3, Cz, P8, C4, and Fz. The right panel shows channel$\times$feature heatmap revealing interactions between electrode locations and feature types in model predictions.

\begin{figure}[!htbp]
\centering
\includegraphics[width=0.92\linewidth]{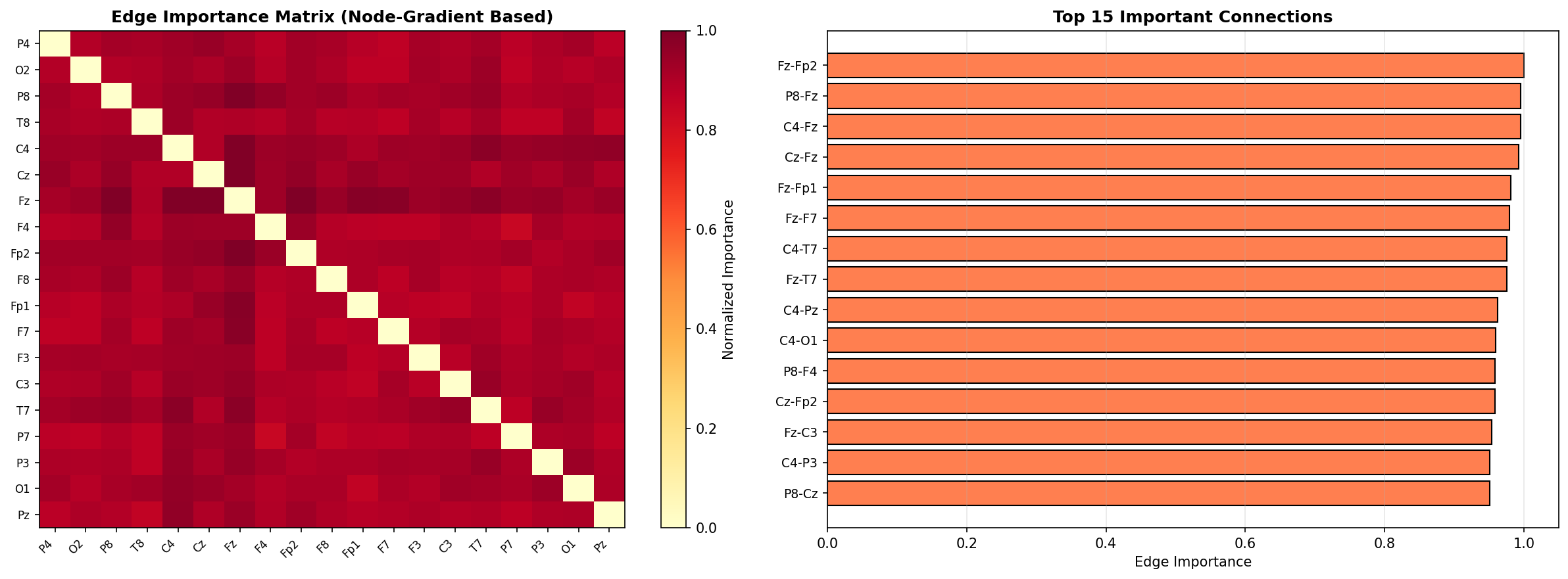}
\vspace{-2mm}
\caption{Edge importance matrix (left) and top 15 important connections (right)}
\label{fig:edge_importance}
\end{figure}

Figure~\ref{fig:edge_importance} shows the edge importance matrix displayed in two panels. The left panel shows the 19$\times$19 importance matrix with cream-red color scale (0.0-1.0), where darker colors indicate higher importance. The right panel displays a bar chart of 15 most important connections (all approaching importance 1.0): Fz-Fp2, P8-Fz, C4-Fz, Cz-Fz, Fz-Fp1, Fz-F7, C4-T7, Fz-T7, C4-Pz, C4-O1, P8-F4, Cz-Fp2, Fz-C3, C4-P3, and P8-Cz. Dominance of connections involving Fz confirms the central role of frontal-midline cortex in classification. This Integrated Gradients-based analysis approach aligns with explainability methodology developed by \citet{sujatha2023} for EEG model interpretation.

\begin{figure}[!htbp]
\centering
\includegraphics[width=0.72\linewidth]{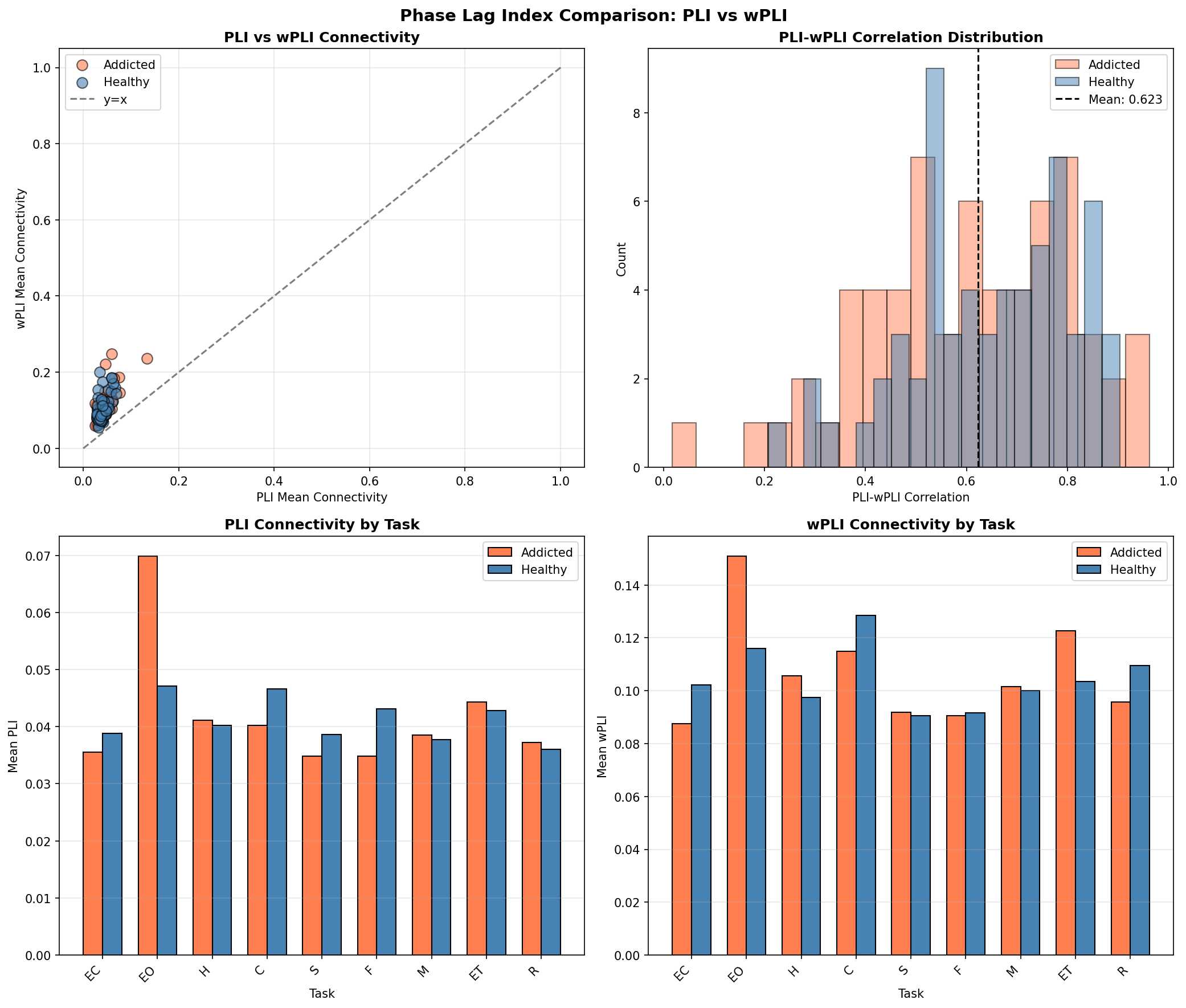}
\vspace{-2mm}
\caption{PLI vs wPLI comparison: scatter plot, correlation distribution, and connectivity per task}
\label{fig:pli_wpli}
\end{figure}

Figure~\ref{fig:pli_wpli} presents PLI and wPLI comparison showing average correlation of 0.623, with wPLI producing 2.5$\times$ higher connectivity values. wPLI is more robust to noise at 0° and 180° phase differences, making it a better choice for EEG connectivity analysis with high volume conduction potential.

\FloatBarrier

The results of this research have significant implications. First, DST-GNN provides an objective neuroscience-based screening tool that can complement self-report methods. High recall (85.71\%) ensures minimization of false negatives. Second, identification of trait-level biomarkers (Cz-T7 connectivity) opens the possibility of detection without pornographic stimulus exposure. Third, identification of frontal-central regions can inform development of more targeted neurofeedback interventions. However, this research still has several limitations. First, small sample size (N=14) limits generalizability and causes high variability. Second, the dataset comes from a single center so cross-cultural validation is needed. Third, age range is limited (13-15 years). Fourth, the model has not been validated on real-time data.

\section{Conclusion}

This study successfully developed a Dynamic Spatio-Temporal Graph Neural Network (DST-GNN) for detecting pornography addiction in adolescents based on EEG signals. DST-GNN achieves F1-Score of 71.00\% $\pm$ 12.10\% with recall 85.71\%, a 104\% improvement compared to the best baseline. Ablation study confirms temporal contribution (21\%) and PLI graph construction (57\%). Frontal-central regions (Fz, Cz, C3, C4) are identified as most discriminative with Beta contribution 58.9\% and Hjorth 31.2\%. Cz-T7 connectivity shows consistent patterns as a trait-level biomarker. These results contribute to child protection efforts by providing an objective screening tool. Future research is suggested to validate on larger datasets with Indonesian population, develop real-time systems, and integrate with neurofeedback protocols.

\section*{Acknowledgments}
The authors would like to express their gratitude to Dr. Kang and the team at Taylor's University for making the public dataset available via Mendeley Data. We also extend our appreciation to Yayasan Kita dan Buah Hati (YKBH) Jakarta for their contributions to the data collection process. This research was conducted in support of the mandate of Law No. 35 of 2014 regarding Child Protection.

\section*{Author Contributions}
\textbf{Achmad Ardani Prasha}: Conceptualization, Methodology, Software, Validation, Formal analysis, Investigation, Data curation, Writing - original draft, Visualization. \textbf{Clavino Ourizqi Rachmadi}: Methodology, Validation, Formal analysis, Writing - review \& editing. \textbf{Sabrina Laila Mutiara}: Methodology, Validation, Writing - review \& editing. \textbf{Hilman Syachr Ramadhan}: Validation, Writing - review \& editing. \textbf{Chareyl Reinalyta Borneo}: Validation, Writing - review \& editing. \textbf{Saruni Dwiasnati}: Supervision, Project administration, Writing - review \& editing.

\clearpage
\bibliographystyle{plainnat}

\end{document}